*Article*

# Toward Sensor-based Sleep Monitoring with Electrodermal Activity Measures

*sensors*

**William Romine [1], Tanvi Banerjee [2] \*, and Garrett Goodman [2]**

[1] *Department of Biological Sciences, Wright State University*; romine.william@gmail.com
[2] *Department of Computer Science and Engineering, Wright State University*; tanvi@knoesis.org
\* Correspondence: tanvi@knoesis.org;

**Abstract:** We use self-report and electrodermal activity (EDA) wearable sensor data from 77 nights of sleep on six participants to test the efficacy of EDA data for sleep monitoring. We used factor analysis to find latent factors in the EDA data, and causal model search to find the most probable graphical model accounting for self-reported sleep efficiency (SE), sleep quality (SQ), and the latent EDA factors. Structural equation modeling was used to confirm fit of the extracted graph. Based on the generated graph, logistic regression and naïve Bayes models were used to test the efficacy of the EDA data in predicting SE and SQ. Six EDA features extracted from the total signal over a night's sleep could be explained by two latent factors, EDA Magnitude and EDA Storms. EDA Magnitude performed as a strong predictor for SE to aid detection of substantial changes in time asleep. The performance of EDA Magnitude and SE in classifying SQ showed promise for wearable sleep monitoring applications. However, our data suggest that obtaining a more accurate sensor-based measure of SE will be necessary before smaller changes in SQ can be detected from EDA sensor data alone.

**1. Introduction**

Sleep is a necessary component to an individual's well-being. Sleep deprivation can cause an increase in day-to-day stress and induce negative emotional responses to routine daily stressors [1,3]. In the long term, continual sleep deprivation can lead to an impaired immune system and increased susceptibility to both chronic and infectious disease [2]. To address these health problems caused by sleep deficiency, we need novel sensing methods for monitoring sleep non-invasively. Traditional measures of sleep quality are self-reported [11]; however, use of sensors to monitor sleep has received increasing attention [12-15]. To this end, electrodermal activity (EDA) may be particularly useful in qualifying sleep activity [20]. In this study, we focus on better understanding the efficacy of EDA data from a wearable sensor called the Empatica E4 (E4) [10] and compare these to self-reported measures of sleep quality (SQ) and sleep efficiency (SE). We focus on better understanding: (1) latencies within features extracted from the EDA signal; (2) the structure of the relationships between the EDA features, SQ, and SE; and (3) the efficacy of the EDA signal in predicting self-reported SQ and SE.

*1.1 Sleep Monitoring Using Surveys*

One of the most well-known self-report sleep surveys is the Pittsburgh Sleep Quality Inventory (PSQI) [11]. The PSQI is a self-report survey using questions to calculate SE, sleep disturbances, and daily sleepiness. The overwhelming availability of fitness sensors and smartphone technology allows us to move beyond self-report data. In Lane et al., the authors designed a smartphone app called "BeWell" to monitor a user's physical activity, social interactions, and sleep patterns [12]. This is done using the smartphone's accelerometer while the device is on the bed where the user sleeps. Similarly, Chen et al. discusses a Best Effort Sleep model which is implemented in the "BeWell" smartphone app previously mentioned [13]. Another model devised by Hao et al. proposes an app called "iSleep" [14]. This system uses the microphone of a smartphone to monitor the factors of sleep such as snoring and body movement to measure the SQ of a user.

As mentioned earlier, wearable sensors are becoming ubiquitous. Sensors like Fitbit [17] and Jawbone [18] are popular and readily available commercial sensors. Muaremi et al. demonstrated sleep monitoring in pilgrims during the Hajj 2013 to detect stress [19]. The authors used two devices, the Zephyr Bioharness 3, and the Empatica E3 which is the previous iteration of the E4 device used in our study. They created a simple two question daily stress questionnaire to monitor sleep and predict stress levels of the pilgrims monitored in the study.

While previous work demonstrated the different approaches of monitoring sleep and making SQ predictions, only the study by Muaremi et al. [19] used a wearable device in conjunction with a questionnaire for sleep monitoring. In the study by Beecroft et al. [16], using only actigraphy or a questionnaire was found to be insufficient for sleep assessment. However, a key difference with [19] and our proposed study is that the authors were predicting stress measures rather than specific sleep measures such as SE or SQ. Also, the studies using only the mobile devices, accelerometer and microphone did not use the physiological signals provided by a wearable sensor. We seek to understand the efficacy of EDA data collected from the non-intrusive wearable E4 device in predicting SE and SQ derived from the daily PSQI (in Supplementary Materials).

*1.2 Sleep Monitoring Using A Controlled Environment*

The most accurate way of monitoring sleep is still done in an enclosed environment using invasive technologies such as the polysomnography (PSG) test. The PSG is an intrusive test requiring several surface electrodes to be placed on the head to evaluate the brain activity, battery packs or extensive wires that can restrict movement in sleep and affect the sleep of the user. The study performed by Douglas et al. [15] used the knowledge of the high performing PSG test and compared other possible methods of sleep monitoring to recommend different approaches to classifying sleep apnea/hypopnea syndrome (SAHS). In Beecroft et al., researchers looked at sleep loss in clinically ill patients and compared PSG results to data gathered from actigraphy and reports from nurses [16]. They concluded that while sleep was severely disrupted in these clinically ill patients, actigraphy and nurse data records were not sufficient to detect loss of sleep. This finding supports the need for non-invasive sensors that can monitor sleep without disrupting normal sleep behavior.

**2. Materials and Methods**

*2.1 Data Collection*

We used the E4 and the daily-PSQI to gather data for a total of 77 nights of sleep from six participants ages 24 to 36, with three male and three female participants. One participant provided a majority of the collected nights of sleep totaling 38 nights of data, similar to the study performed by Sano et al. [20]. Of the remaining five participants, one participant provided four nights of data, and the remaining four participants provided seven nights each. During the data gathering process, the participants all wore the device to bed at night. When they got up, they would remove the device and answer the questions of the daily-PSQI using an Android application.

*2.2 Instrumentation*

The device used in this study, the Empatica E4, is an unobtrusive wrist-worn physiological sensor. The E4 is similar to a Fitbit in design and size [17] and is capable of reading Blood Volume Pulse (BVP), Heart Rate (HR), Interbeat Intervals (IBI), Skin Temperature (TEMP), 3-Axis Accelerometer (ACC), and Electrodermal Activity (EDA) [10]. We focus on the EDA feature which is the measurement of electrical conductance of the skin for detection of sympathetic nervous system arousal [21].

From the daily-PSQI, we calculated sleep efficiency (SE) as a function of the number of minutes spent asleep divided by the number of minutes spent in bed as reported by the participants. The ratings from the SQ range from one as very poor and four as very good. We defined ratings of 3-4 as "good sleep" and 1-2 as "poor sleep".

*2.3 Creating Models for Sleep Efficiency (SE) and Sleep Quality (SQ)*

2.3.1 Feature Extraction

We extracted 8 EDA features (Figure 1) using the automatic feature extraction method described by Sano et al. [20]. We calculated the awake segments for derivation of SE using the E4's 3-axis accelerometer and the function described by Cole et al. [22] but found that these did not correlate well with self-report measures (r = 0.17). Given this low correlation and that Cole et al.'s derivation has not been validated clinically, we proceeded to use the self-report derivation for SE in this study.

**Table 1.** EDA features extracted from the Empatica E4 and self-report features drawn from the daily-PSQI.

| *Feature* | *Type* | *Description* |
|---|---|---|
| Amount of Sleep Minutes | PSQI Feature | The number of minutes the participant was asleep. |
| Amount of Wake Minutes | PSQI Feature | The number of minutes the participant was awake. |
| Number of EDA-Peak Epochs | EDA Feature | The number of EDA-peak epochs contained in the EDA signal. |
| Number of EDA Storms | EDA Feature | The number of EDA storms contained in the EDA signal. |
| Average Size of EDA Storms | EDA Feature | Mean of all the EDA storms of a single night. |
| Standard Deviation of EDA Storms | EDA Feature | Standard deviation of all the EDA storms of a single night. |
| Largest EDA Storm | EDA Feature | Largest EDA storm in the signal for a single night. |
| Number of EDA Events | EDA Feature | Number of EDA events (peaks) in the signal for a single night. |
| SE | PSQI Feature | SE calculated using the daily-PSQI questions 1, 2, and 3. |
| SQ | PSQI Feature | SQ deduced using SQ rating scale from the PSQI and creating a binary system where bad is 1-2 rating and good is 3-4 rating. |

2.3.1 Model Selection

Obtaining a proper understanding of how PSQI and E4 sensor readings relate to each other required specification of a directed acyclic graph (DAG) that captures accurately the causal connections between these variables. DAGs allow derivation of causal connections between self-report and sensor sleep data without confounding factors and type 1 error inflation which often persist in ordinary linear stepwise regression methods [4]. In light of the problem described, our inferential process was fourfold: (1) search for latent factors in the extracted E4 features, (2) learn the structure of the causal connections (the DAG) between the observed or latent variables, (3) verify that this DAG reproduces the data adequately, and (4) use the DAG to derive statistical inferences which align with our research questions.

*Latent Variable Search:* We used exploratory factor analysis (EFA) with maximum likelihood estimation and promax rotation [5] to search for latent factors extracted from the EDA signal. We retained all factors with a variance greater than that provided by a single variable (i.e. eigenvalue > 1) [6]. Regression coefficients of the promax-rotated latent factor scores on the features were used to define the meaning of the latent factors qualitatively.

*Learn the DAG from the Data:* We then utilized causal model search to derive a DAG explaining causal links between PSQI and E4 observed or latent features that are best supported by our data. We used the Fast Greedy Search (FGS) algorithm within the TETRAD software package [7] to extract the most likely DAG given the data using the features from the EDA signal, sleep quality, and sleep efficiency. As a score-based approach, the FGS selects the most probable model given the data. We used the Conditional Gaussian Bayesian Information Criterion (BIC) score with a structure prior of 1 [8].

*Verify that the DAG Explains the Data:* We then used a structural equation modeling (SEM) framework to evaluate statistically the extent to which the DAG explains the data as well as the strength of edges in the model [9]. We used Pearson's chi-square, the root mean square error of approximation (RMSEA) and comparative fit index (CFI) as measures of fit. RMSEA is a measure of absolute fit, with measures of 0.07 or below indicating adequate fit [23]. The CFI is a measure of relative fit with respect to the independence model, where a value of 1 indicates perfect fit, and a value of 0 indicates a level of fit no better than that offered by treating the features as independent. CFI values of 0.90 or above indicate adequate fit [24]. Estimation of path coefficients was done using the diagonally weighted least squares (WLSMV) estimator in Mplus 7 [25].

We selected the set of relevant predictors by taking the Markov blanket for the nodes of sleep efficiency and sleep quality, respectively, on the DAG [26]. Upon selecting the features based on the Markov blanket, ordinary least squares linear regression was used to evaluate the efficacy of the selected sensor features ($\alpha$ = 0.05 level) in predicting SE, and binary logistic regression was used to evaluate SQ ($\alpha$ = 0.05 level). In addition, we used the Naïve Bayes classifier as a generative counterpart to gain a more complete picture of how well SE and the EDA features extracted from the E4 sensor predict SQ.

## 3. Results

The six EDA features could be described adequately with two latent factors. The first factor contained 4.5 variables of variance, accounting for 75% of the total variation in the set of variables. Five features (number of EDA epochs, size of the EDA event, standard deviation of the EDA event, the largest EDA storm, and the number of EDA events) had high loadings (0.87-1) onto this first factor, suggesting that these all describe a single latent factor: magnitude of EDA activity through the night, which we call "EDA Magnitude" for the remainder of the paper. Number of EDA storms had a small loading (-0.09) onto the EDA Magnitude factor, but instead loaded highly (0.83) onto the second factor, Number of EDA storms, which we call "EDA Storms" from here on. This second factor contained 1.2 variables of variance for an additional 20% of the total variance in the data. In total, the 2-factor solution described 95% of the variance in the six EDA features.

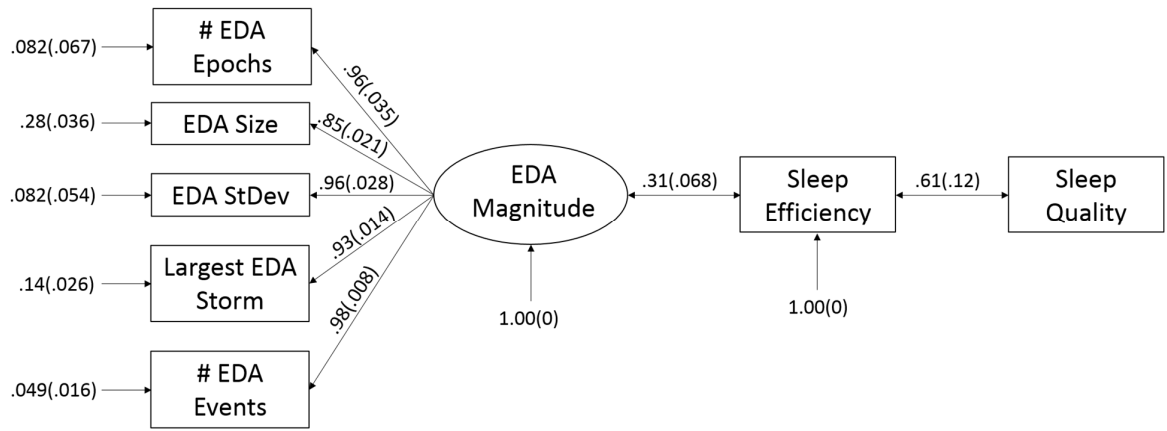

**Figure 1.** Path diagram of the relationship between EDA features extracted from the E4 sensor, reported sleep efficiency, and reported sleep quality. Standardized coefficients are reported. All are significant at the 0.05 level.

The FGS yielded the model displayed in Figure 1. SEM suggested that this model replicated the covariance between the observed variables well (RMSEA = 0.064, CFI = 0.97) and could not be rejected by the data ($\chi^2_{df=22}$ = 28.69, p = 0.15). Paths from the latent variable of EDA Magnitude to the observed sensor features were strong (0.85-0.98) corroborating the EFA that these measure a single latent factor. We found a significant connection between EDA Magnitude and reported SE (b = 0.31, SE = 0.68, z = 4.6, p << 0.001), and between reported SE and SQ (b = 0.61, SE = 0.12, z = 5.3, p << 0.001). This indicates that the features extracted from the E4 sensor may be useful in predicting both SE and SQ.

The Markov blanket for SE on the DAG in Figure 1 suggests that EDA Magnitude and SQ can be used as direct predictors of SE. The Markov blanket on SQ suggests that SE is sufficient to predict SQ without need for EDA data, but that EDA features may also be useful for prediction of SQ through their ability to predict SE.

The significant relationship between EDA Magnitude extracted from the E4 sensor and SE after controlling for reported SQ ($r_{partial}$ = 0.28, p = 0.018) provides support for the utility of EDA sensor data in addition to self-report data for sleep monitoring applications. A linear regression model using only the sensor data with 10-fold cross-validation fits with a root mean square error (RMSE) of 0.99, a mean absolute error (MAE) of 0.75. This model is statistically significant ($F_{df=1,73}$ = 7.59, p = 0.007, $r^2_{adj}$ = 0.082, $SE_{reg}$ = 0.96). This indicates that using only the sensor data without self-reported information may provide a promising model to predict SE.

Binary logistic regression using both EDA Magnitude and SQ as predictors ($\chi^2_{df=2}$ = 16.4, p << 0.001, Cox & Snell $r^2$ = 0.20) demonstrates the need for including SE as a predictor for SQ. After accounting for SE, a participant's EDA activity has no significant effect on SQ (OR = 1.2, $\chi^2_{df=1}$ = 0.31, p = 0.58). SE is useful for predicting SQ (OR = 3.7, $\chi^2_{df=1}$ = 10.6, p = 0.001). The logistic regression model reported here predicts SQ with an AUC of 0.76 and an F1 score of 0.73 (precision = 0.73, recall = 0.76). The Naïve Bayes classifier with these features also performs well, with an AUC of 0.85 and an F1 score of 0.83 (precision = 0.82, recall = 0.83). When only SE is used as a predictor of SQ (without the EDA sensor data), the predictive power of the naïve Bayes model decreases slightly: AUC of 0.83 and F1 of 0.80 (precision = 0.80, recall = 0.80). The predictive power of the logistic regression model remains unchanged, with an AUC of 0.77 and an F1 of 0.73 (precision = 0.73, recall = 0.76). However, given our relatively small number of training samples relative to many machine learning applications, generative models approach their asymptotic error more quickly than discriminative models, explaining why naïve Bayes outperforms logistic regression in this study and benefits from EDA magnitude being added in addition to SE [27].

**4. Discussion**

Our data suggest that EDA measures show promise toward sleep monitoring, but more work remains for a deeper understanding of physiological changes occurring during the different sleep stages (such as the differences in REM vs SWS (slow wave sleep)). After controlling for reported SQ, EDA magnitude had a **significant** relationship with SE, supporting the usefulness of EDA measures toward predicting SE. This indicates that using only the sensor data without self-reported information may be useful in measuring SE. However, since the EDA features predict SE with an RMSE of 0.99 and MAE of 0.75, our data suggest that only a coarse (high/low) measure of SE can be attained through the use of sensor data alone. This implies that a sensor using these EDA features may be useful for detecting large anomalies in time asleep which may be induced by changes in medication, major lifestyle changes, or large changes in living environment. That said, more work is needed before such a device could be used to detect fine deviations in SE. This would indicate the use of additional features such as heart rate variability, along with the current EDA measures.

Figure 1 suggests that SE is a mediator between measures of EDA magnitude and SQ; however, insufficient evidence is available to tease out the directionality of these relationships i.e. whether EDA causes SQ or changes in SE and SQ result in EDA changes. It suffices to say, however, that when SE is accounted for, the relationship between EDA magnitude and SQ is non-significant. It is not surprising, then, that omitting the sensor data and just using SE as a predictor for SQ yields no significant difference in the classification performance when logistic regression is used. Naïve Bayes, however, tells a slightly different story, generating a small but noticeable deterioration in classification performance (F1 score decreasing from 0.83 to 0.80) when the EDA magnitude is removed from the model. This indicates that having both EDA data and a measure for SE may be helpful for a higher performance when a generative modeling framework is used. This is an important distinction since currently we have 77 data samples, so the generative model is supposed to have lower asymptotic error [CITE NG] and including the simple EDA features does not add much to the computational complexity of the model while slightly improving model performance. Whether a discriminative or generative modeling approach is used, a useful classifier for high vs low SQ is obtainable so long as an accurate and repeatable measure of SE exists.

Toward understanding the efficacy of EDA data in informing us about SE and SQ on a nightly basis, a reader may ask why we did not use a sensor-related measure of SE such as that described in [22]. We had considered using Cole's function [22] for calculation of SE that utilized the accelerometer data from the X, Y, and Z directions. However, this showed poor correlation with the self-reported SE values used in this study (r=0.17, $F_{df=1,73}$ = 2.03, p = 0.16), meaning that they only shared ~3% of their variance in common. We consider self-reported SE as a ground truth, so given this negligible relationship, we decided not to use this sensor-derived measure in our current modeling. This computation using the accelerometer was reported in the study by Sano et al. [20] but not validated (they used an earlier version of the E4 device). Possible reasons for the poor performance using the movement (in x, y, and z directions) alone highlights the differences in individual sleep habits such certain participants being more active in their sleep as compared to others, which a reliable automated sensor must be able to capture. Clearly, more customized methods are needed to cater Cole's function to the habits of specific individuals.

## 5. Conclusions

Our data indicate that a combination of sensor data and self-reported measures can be used to generate a useful metric to both quantify and qualify sleep. However, we still have a long way to go towards using only sensor data, especially while measuring the quality of sleep. That said, we have taken a promising first step towards understanding how to quantify sleep measures (SE and SQ) using EDA sensor data. We hope this will be used for continuous assessment of an individual's sleep behavior and potentially record large changes or deteriorations in their sleep patterns over time. The next step involves including more physiological features to measure SE and SQ to enable detection of smaller changes over time and capture more subtle aspects of sleep behavior over time.

**Supplementary Materials:** our PSQI.

**Author Contributions:** WR conducted the data analysis and drafted the Introduction, Methods, Results, and Discussion sections. TB conceptualized the study and drafted the Literature Review and Discussion sections. GG collected the data, extracted the features, and contributed to the Literature Review.

**Funding:** This work was supported in part by the NIH under grant K01 LM012439-02.

**Acknowledgments:** The authors thank the participants who made this study possible.

**Conflicts of Interest:** The authors declare no conflict of interest.